\documentclass[10pt,twocolumn,letterpaper]{article}

\usepackage[pagenumbers]{cvpr} %

\usepackage{lipsum}

\definecolor{cvprblue}{rgb}{0.21,0.49,0.74}
\usepackage[pagebackref,breaklinks,colorlinks,allcolors=cvprblue]{hyperref}
\usepackage{cuted}
\usepackage{gensymb}
\def\confName{CVPR}
\def\confYear{2026}

\title{Image-Guided Geometric Stylization of 3D Meshes}

\author{
    Changwoon Choi$^{*1}$
    \hspace{5mm}
    Hyunsoo Lee$^{*1}$
    \hspace{5mm}
    Cl\'ement Jambon$^2$
    \hspace{5mm}
    Yael Vinker$^2$
    \hspace{5mm}
    Young Min Kim$^{\dagger1}$\\
    {
    $^1$Seoul National University
    $\quad$$^2$Massachusetts Institute of Technology (MIT)
    }
}
\newcommand{\bx}{\mathbf{x}}

\newcommand{\bc}{\mathbf{c}}

\begin{document}
\twocolumn[{
\renewcommand\twocolumn[1][]{#1}
\maketitle
\vspace{-1.6em}
\begin{center}
    \centering
        \captionsetup{type=figure}
        \includegraphics[width=\linewidth]{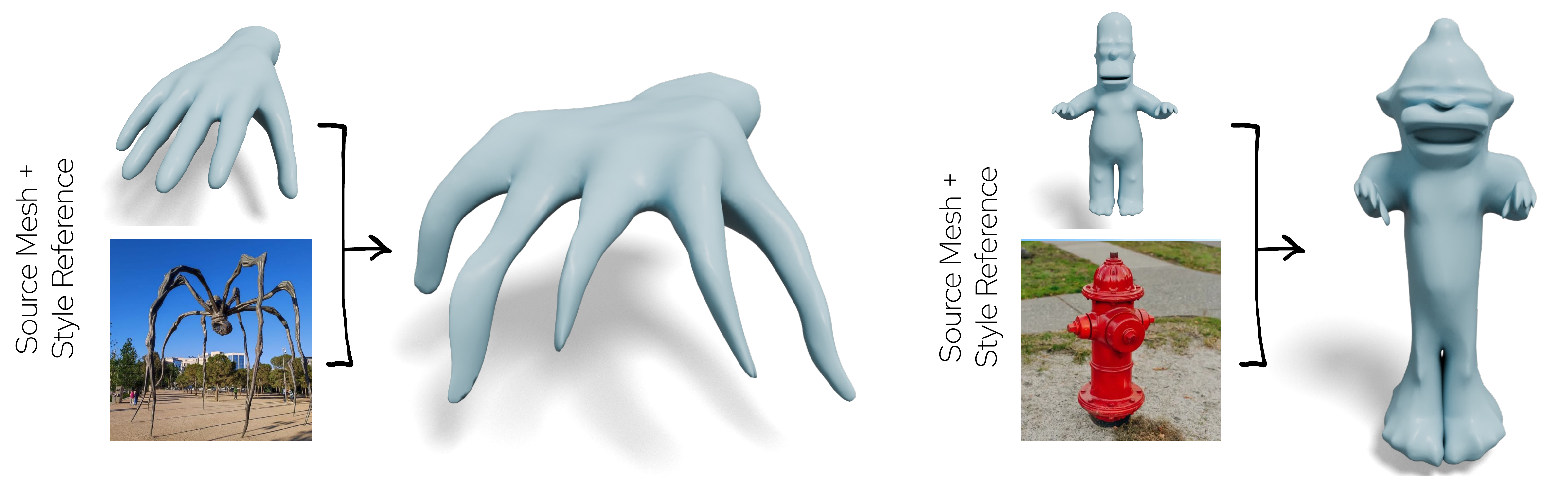}
        \captionof{figure}{
        Our method enables stylization of 3D meshes driven by image style references.
        The target stylized meshes retain the coarse structure and semantics of the input mesh while incorporating internal geometry derived from the style reference.
        }
        \label{fig:teaser}
\end{center}
}]
\maketitle
\def\thefootnote{*}\footnotetext{Equal contribution.}
\def\thefootnote{$\dagger$}\footnotetext{Young Min Kim is the corresponding author.}
\def\thefootnote{\arabic{footnote}}

\begin{abstract}
Recent generative models can create visually plausible 3D representations of objects.
However, the generation process often allows for implicit control signals, such as contextual descriptions, and rarely supports bold geometric distortions beyond existing data distributions.
We propose a geometric stylization framework that deforms a 3D mesh, allowing it to express the style of an image. 
While style is inherently ambiguous, we utilize pre-trained diffusion models to extract an abstract representation of the provided image.
Our coarse-to-fine stylization pipeline can drastically deform the input 3D model to express a diverse range of geometric variations while retaining the valid topology of the original mesh and part-level semantics.
We also propose an approximate VAE encoder that provides efficient and reliable gradients from mesh renderings.
Extensive experiments demonstrate that our method can create stylized 3D meshes that reflect unique geometric features of the pictured assets, such as expressive poses and silhouettes, thereby supporting the creation of distinctive artistic 3D creations.
Project page: \url{https://changwoonchoi.github.io/GeoStyle}

\end{abstract}
 
\vspace{-1.4em}
\section{Introduction}
\label{sec:intro}

Despite impressive advancements in 3D generative models, it is not trivial to support artistic creation of 3D models with existing AI tools.
The artistic style is beyond what one can achieve by interpolating or composing existing data distributions, and the unique component embedded within the imagined creation cannot be communicated intuitively.
Previous attempts at stylization focus on altering the local appearances of the given global structure.
Image stylization works distinguish style from content, maintaining the overall layout of the original image while altering only local patch statistics~\cite{gatys2016image}.
Similarly, 3D stylization methods incorporate text descriptions to guide local geometric variations on the surface~\cite{michel2022text2mesh,dinh2025geometry}, or produce specific geometric characteristics with handcrafted regularizations~\cite{liu2021normal,cubic_stylization,gauss_stylization}.

\begin{figure}[t!]
	\centering
	\includegraphics[width=\linewidth]{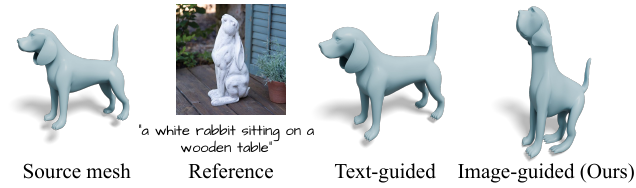}
	\caption{
    Text-based deformation often struggles to capture and transfer geometric style, whereas our image-guided framework can transfer intended aesthetic from rich and specific visual cues.
    }
    \label{fig:comparison_text_and_image_reference}
\end{figure}

We expand the notion of style beyond high-frequency textures to embrace geometric features of various scales as components of a unique style.
For example, in~\cref{fig:teaser}, the distinctive silhouette of Bourgeois's spider or the rigid structural characteristics of a fire hydrant cannot be described by local texture.
Such a diverse range of variations requires a holistic analysis, whose geometric characteristics are challenging to describe unambiguously with an input text, as shown in~\cref{fig:comparison_text_and_image_reference}.
We utilize reference images as a means of explicit description to inspire 3D stylization intended by users.
With the power of image diffusion models, the image clue can be transformed into a style-specific optimizer for free, which can guide geometric stylization.

Instead of generating stylized geometry from scratch, we formulate geometric stylization as the task of deforming a user-provided mesh model.
The deformation maintains the original manifold topology despite a significant change in structure.
While the popular choice of volumetric representations can generate visually plausible 3D shapes with a differentiable rendering pipeline, these representations often lack a rigorous topological structure.
In contrast, we can start from a valid mesh topology and maintain compatibility with valuable assets, such as UV maps and motion rigs, as well as rich geometric processing pipelines for smoothing, upsampling, and re-meshing.

Our stylization framework requires the extraction of geometric style and significant yet valid deformation of the given mesh.
We define the stylistic component as an abstract feature of a pre-trained large-scale diffusion model~\cite{podell2023sdxl} and extract the style of the reference image as LoRA weights~\cite{hu2022lora}.
Then, Score Distillation Sampling (SDS)~\cite{poole2022dreamfusion} drives mesh deformation to align with the reference style.
We propose using an approximate VAE encoder of the latent-based diffusion model, which is crucial for stabilizing the optimization in practice. 
Our deformation pipeline first encourages semantically coherent deformation per part at a coarse level, followed by finer deviation via Jacobian optimization~\cite{aigerman2022neural}.
We optionally preserve symmetry, which maintains internal consistency.
Together, these components form a general and practical framework for transferring high-level geometric style from 2D images to 3D meshes, paving the way for intuitive, reference-driven 3D content creation.

\section{Related Work}
\label{sec:related_works}

\paragraph{Style transfer.}
Style transfer is the task of applying the visual style of one input to the content of another, producing a stylized output that preserves the original content while adopting the target style.
The seminal work by Gatys et al.~\cite{gatys2016image} and its follow-ups~\cite{chen2016fast,kolkin2019style,li2016combining,gu2018arbitrary,liao2017visual,mechrez2018contextual,risser2017stable,kotovenko2019content,kotovenko2019content_cvpr,kotovenko2021rethinking} formulated style transfer as an iterative optimization problem that minimizes content and style losses defined on deep features~\cite{vggnet}.
Another line of works~\cite{huang2017arbitrary,an2021artflow,park2019arbitrary,sheng2018avatar,li2017universal} employs feed-forward networks that learn to approximate the optimization-based process with a neural network, enabling efficient stylization. 
While these approaches show promising results, they primarily focus on matching patch statistics and transferring high-frequency appearance attributes such as color, texture and brushstroke patterns.

Image style transfer has been naturally extended to 3D style transfer, aiming to stylize a 3D scene so that its rendered appearance matches the visual characteristics of a reference image.
Recent works have explored style transfer across various 3D representations, including point clouds~\cite{mu20223d,huang2021learning}, meshes~\cite{hollein2022stylemesh}, Neural Radiance Fields~\cite{zhang2022arf,zhang2023ref,pang2023locally,chiang2022stylizing,jung2024geometry} and 3D Gaussian Splatting~\cite{liu2024stylegaussian,zhang2024stylizedgs}.
Similar to image style transfer, most 3D style transfer approaches mainly focus on transferring high-frequency surface textures while overlooking the role of geometry in conveying style.
Only a few works have explored geometric style transfer; however, they are restricted to the image domain~\cite{liu2020geometric,yaniv2019face} or limited to specific object categories~\cite{yin20213dstylenet,yaniv2019face}.

\paragraph{3D mesh stylization by deformation.}
In contrast to texture-based style transfer on 3D meshes, some approaches stylize meshes by deforming them.
Early works~\cite{liu2018paparazzi,kato2018neural} and its follow-ups~\cite{dynamic_mesh_stylization} optimize positions of mesh vertices by minimizing image-space style loss~\cite{gatys2016image,kolkin2022neural} with differentiable rendering technique.
Hertz et al.~\cite{geometric_texture} transfer geometric textures from reference meshes to source meshes.
However, they are limited to transferring high-frequency geometric details and only achieve local vertex displacements.
Another line of work~\cite{cubic_stylization,gauss_stylization,liu2021normal} stylize meshes with large-scale deformation by minimizing hand-crafted heuristic regularization terms, which are limited to representing specific styles.
Recent works~\cite{gao2023textdeformer,kim2025meshup,dinh2025geometry} utilize large image models, including CLIP~\cite{radford2021learning} and pixel-space diffusion model~\cite{DeepFloydIF}, for mesh deformation.
By leveraging powerful large image models which have high-level and semantic understanding of shapes, the methods successfully deform the source mesh into various concepts or styles.
However, the inherent ambiguity of text descriptions often makes it difficult to precisely control the stylization or reflect the user’s stylistic intent.

\section{Method}
\label{sec:method}

\begin{figure}[t!]
	\centering
	\includegraphics[width=1.0\linewidth]{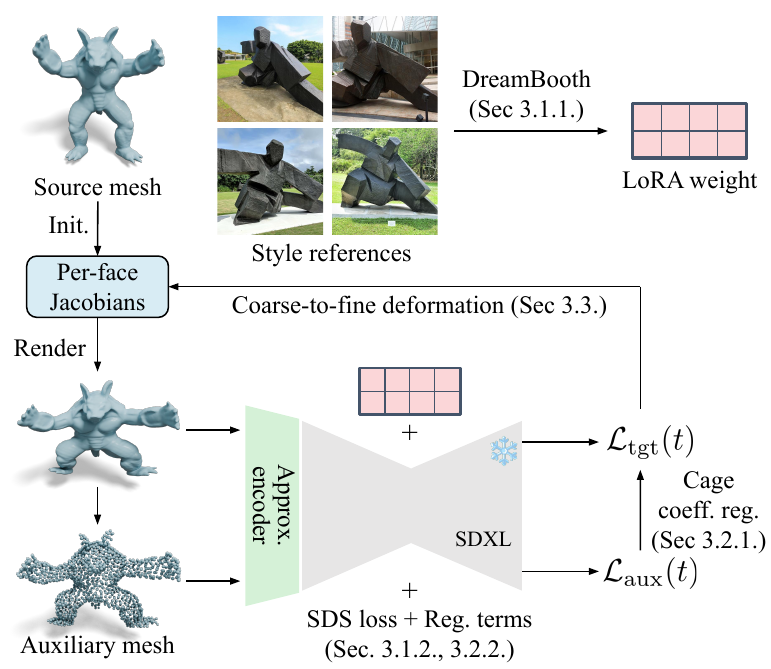}
	\caption{
    Method overview.
    We first extract geometric style from reference images by training LoRA using a DreamBooth-style objective (\cref{subsec:dreambooth}). 
    We optimize per-face Jacobians to transfer geometric style to source mesh. 
    The optimization is guided by SDS loss (\cref{subsec:sds_loss}).
    To handle both large structural deformation and fine-grained details, we adapt a coarse-to-fine deformation (\cref{subsec:deformation}), along with the cage-based regularization (\cref{subsec:cage_reg}) and optional symmetry regularization (\cref{subsec:sym_loss}).
    }  
\label{fig:method}
\end{figure}

Given a source mesh and reference style images, our method deforms the mesh to exhibit the geometric style depicted in the input images. 
An overview of our pipeline is illustrated in~\cref{fig:method}.
We first extract the abstracted style component from input images as a LoRA weight of latent diffusion model.
Then we deform the mesh by minimizing the SDS loss from a style-infused latent diffusion model with an efficient low-rank approximation of the encoder.
The deformation adapts our novel coarse-to-fine strategy, which effectively preserves content identity and local features without deteriorating the mesh structure.
In addition to per-face optimization of the mesh to minimize the proposed loss, we introduce an auxiliary representation with coarse samples, and allow large-scale changes with part-level regularization.

\subsection{Style Matching}

The primary subtask in geometric stylization is to extract the \textit{style} encoded in the reference images and to optimize the source mesh.
Although ‘style’ is difficult to define explicitly, we leverage the generative priors of pre-trained diffusion models to extract a coherent abstraction of it.

\subsubsection{Image Style Extraction}
\label{subsec:dreambooth}

DreamBooth~\cite{ruiz2023dreambooth} provides a critical mechanism for this -- by fine-tuning a pretrained diffusion model on a small set of reference images, it encourages the model to capture the distinctive attributes that define the subject's appearance:
\begin{equation}
    \mathcal{L}_{\mathrm{DreamBooth}} = \mathbb{E}_{\bx, \bc, \mathbf{\epsilon}, t} [w_t \| \hat{\bx}_{\theta}(\alpha_t \bx + \sigma_t \mathbf{\epsilon}, \bc) - \bx \|^2_2 ],
    \label{eq:dreambooth}
\end{equation}
where $t \sim \text{Uniform}([0, 1])$, $\bx$ denotes a reference image, $\bc$ is the conditioning signal such as a text prompt, and $\alpha_t, \sigma_t, w_t$ are the noise scheduling parameters of the diffusion model~\cite{ho2020denoising, song2020denoising}.
Importantly, the attributes captured through this process extend beyond surface-level texture or fine-scale appearance:
DreamBooth has been shown to preserve coherent shape, proportions, and other structural traits that form the global identity of the subject. 
This aligns with our goal for \textit{geometric stylization}, which incorporates both local details and global shape priors.
We further reduce the computational overhead by restricting the parameter updates to low-rank adapters (LoRA)~\cite{hu2022lora} inserted into the diffusion model's U-Net~\cite{ronneberger2015u}. 
Once it is extracted from the input, the compact abstraction serves as the stylization target that the geometry should match.

\subsubsection{SDS Loss with an Approximated VAE Encoder}
\label{subsec:sds_loss}

We deform the input mesh based on the Score Distillation Sampling (SDS) loss that leverages a pre-trained diffusion model~\cite{rombach2022high}.
Following DreamFusion~\cite{poole2022dreamfusion}, the SDS loss is derived by omitting the Jacobian term of the U-Net~\cite{ronneberger2015u} in the gradient of the original diffusion training loss:
\begin{equation}
    \mathcal{L}_{\mathrm{Diff}} = \mathbb{E}_{\bc, \mathbf{\epsilon}, t} [w_t \|  \mathbf{\epsilon}_{\theta}(\mathbf{z}_t, \bc, t) - \mathbf{\epsilon} \|^2_2 ],
    \label{eq:diffusion}
\end{equation}
where $\mathbf{\epsilon} \sim \mathcal{N}(\mathbf{0}, \mathbf{I})$, $\mathbf{\epsilon}_{\theta}$ represents the noise prediction network of the diffusion model, and $\mathbf{z}_t$ is the noised latent of ground-truth data.
This formulation provides an update direction that follows the score function of the diffusion model toward high-density regions of the data distribution, without requiring backpropagation through the pretrained diffusion model.

To deform the source mesh, we adopt the per-face Jacobian parameterization introduced by Neural Jacobian Fields~\cite{aigerman2022neural}.
More concretely, we avoid directly optimizing vertex positions, which has been shown to be unstable and prone to producing local, fragmented deformations~\cite{gao2023textdeformer}.
Instead, we represent the deformation using per-face Jacobians $\mathbf{J}_i \in \mathbb{R}^{3 \times 3}$, which enables more coherent and larger-scale shape changes.
The deformed vertex positions are then recovered from the deformation map $\phi^\ast$ by solving the following Poisson equation:
\begin{equation}
    \phi^* = \min_{\phi} \sum_i |t_i| \| \nabla_i \phi - \mathbf{J}_i  \|^2.
    \label{eq:possion_eq}
\end{equation}
With this parameterization, the gradient of the SDS loss with respect to the per-face Jacobians $\{\mathbf{J}_i\}$ is expressed as
\begin{equation}
    \nabla_{\mathbf{J_i}}\mathcal{L}_{\mathrm{SDS}} = \mathbb{E}_{\bc, \mathbf{\epsilon}, t}  \left[ w_t (  \mathbf{\epsilon}_{\theta}(\mathbf{z}_t, \bc, t) - \mathbf{\epsilon} ) \frac{\partial \mathbf{z}_t}{\partial \mathbf{J}_i} \right],
    \label{eq:sds_loss}
\end{equation}
where $\mathbf{z}_t$ is the noisy latent obtained from the rendered mesh image, computed with per-face Jacobian $\mathbf{J}_i \in \mathbb{R}^{3 \times 3}$ of a triangular mesh.

\begin{figure}[t!]
	\centering
	\includegraphics[width=0.95\linewidth]{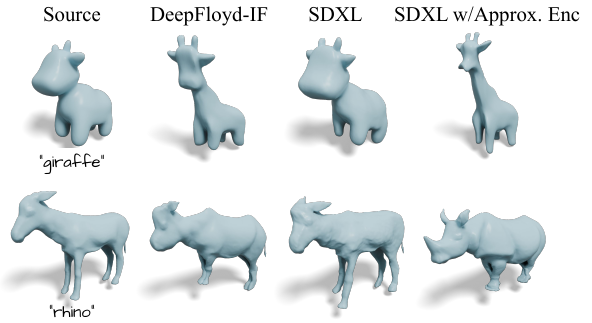}
	\caption{Effect of model choices on text-guided mesh deformation.
    For each source mesh, we apply SDS~\cite{poole2022dreamfusion} guidance using (1) DeepFloyd-IF~\cite{DeepFloydIF}, (2) SDXL~\cite{podell2023sdxl} with the original VAE~\cite{kingma2013auto}, and (3) SDXL with our approximated encoder. 
    Our setup successfully transfers the intended semantic shape, while the alternatives either distort geometry or fail to deform the global structure.
    }
\label{fig:model_cmp}
\end{figure}

\paragraph{Approximated VAE encoder.}

The geometric stylization process extracts features from mesh renderings and propagates gradients to refine the underlying geometry.
Although latent diffusion models such as Stable Diffusion XL (SDXL)~\cite{podell2023sdxl} provide strong generative priors, their large VAE encoder–decoder architecture can make gradient propagation less effective for geometry optimization.
Previous work has shown that approximating components of the VAE in latent diffusion models can significantly accelerate 3D reconstruction tasks~\cite{metzer2023latent}.
Building on this insight, we develop an efficient approximation of the VAE encoder tailored to our setting, enabling fast and stable encoding of rendered images into the latent space during optimization.

Let $\bx \in \mathbb{R}^{3 \times H \times W}$ be a rendered image of the source mesh and $\mathbf{z} \in \mathbb{R}^{4 \times H \times W}$ be its corresponding latent encoded from the SDXL VAE.
We then compute a matrix $\mathbf{A} \in \mathbb{R}^{4 \times 4}$ such that $\mathbf{z} \simeq \mathbf{A} \bar{\bx}$, where $\bar{\bx}= [\bx; \mathbf{1}]$ is obtained by concatenating a one matrix along the channel dimension, allowing to model affine components.
Given $N$ rendered image-latent pairs $\{ (\bx_i, \mathbf{z}_i) \}_{i=1}^N$, we fit $\mathbf{A}^*$ via least squares:
\begin{equation}
    \mathbf{A}^* = \arg \min_{\mathbf{A}} \sum_{i=1}^N \| \mathbf{z}_i - \mathbf{A} \bar{\bx}_i \|^2_2,
    \label{eq:approx_enc}
\end{equation}
where we use $N=500$ in practice.

We emphasize that this approximation strategy is crucial, as even text-guided mesh deformation fails without it.
To justify our choice of using SDXL with an approximated VAE encoder, we compare three setups on text-based mesh deformation:
(1) DeepFloyd-IF~\cite{DeepFloydIF} -- the pixel diffusion model used in MeshUp~\cite{kim2025meshup}, (2) SDXL with its native VAE, and (3) SDXL with our approximated encoder.
As shown in \cref{fig:model_cmp}, the na\"ive SDXL setup struggles to deform the global shape meaningfully, and the DeepFloyd-IF variant often shows suboptimal results.
In contrast, our setup yields semantically aligned deformations, demonstrating its effectiveness. 
Therefore, we adopt SDXL with the approximated encoder for all subsequent experiments.

\subsection{Regularization with Identity Preservation}
\label{subsec:coarse_to_fine}

By allowing deformation to match the style of the input images, we can transform the mesh into the desired style.
While the Jacobian field in Eq.~\eqref{eq:possion_eq} can progressively update the surface details, extreme deformation can deteriorate the overall structure when the reference image is significantly different from the initial geometry, as shown in the second row of~\cref{fig:qual_ablation}.
We propose a coarse-to-fine strategy, enabling large-scale changes at an early stage. 
At a coarse level, we define a cage loss $\mathcal{L}_\text{cage}$, which creates locally coherent structures and preserves the semantics of the geometry (\cref{subsec:cage_reg}).
We assume that the part-level decomposition of the mesh can provide clues for relative semantics, facilitating content preservation, and define coarse-level deformation using cages per part.
Then we gradually increase the relative contribution of the Jacobian optimization  $\mathcal{L}_\text{SDS}$, which refines fine-scale details.
Optionally, we detect the reflective symmetry of the input mesh and preserve it (\cref{subsec:sym_loss}).

\subsubsection{Auxiliary Mesh and Cage-Guided Deformation}
\label{subsec:cage_reg}

In addition to per-face Jacobians, cage-guided deformation coherently moves the large-scale semantic structures.
To disregard the detailed triangulation, we process the coarse-level changes on an auxiliary mesh composed of spheres extracted from the vertex samples of the original mesh.
Additionally, we extract semantic mesh parts using PartField~\cite{liu2025partfield}.
Then we fit Oriented Bounding Boxes (OBBs) aligned with the part segmentation, denoted as $\{ \mathcal{C}_l \}_{l=1}^L$.

The coarse deformation is parameterized by scale $s_l$, rotation $R_l$, and translation $T_l$ of each OBB $C_l$. 
At each optimization step, we first update the OBB parameters $\{ s_l, R_l, T_l \}_{l=1}^L$ using the SDS loss calculated with the rendered view of the auxiliary mesh, denoted as $\mathcal{L}_{\mathrm{SDS}}^{\mathrm{aux}}$. 
The centers of the auxiliary spheres are directly updated by applying the optimized cage transformations.
Concretely, a sphere center with initial coordinate $\mathbf{p} = (x, y, z)$ is translated into $\mathbf{p}' = (x', y', z')$ as $ \mathbf{p}' = s_l \mathbf{R}_l \mathbf{p} + \mathbf{T}_l$.

The part-wise transform is transferred from the auxiliary mesh to the deformation of the source mesh, guided by cages.
We define cage coefficients $\mathbf{W}_i = [w_{i1}, ..., w_{i8}]^{\top}$ that satisfy
\begin{equation}
    \mathbf{v}_i = \sum_{j=1}^8 w_{ij} \mathbf{c}_{lj},
\end{equation}
where these coefficients indicate how much influence the $j$-th OBB corner $\mathbf{c}_{lj}$ has on the position of mesh vertex $\mathbf{v}_i$.
Note that they satisfy $\sum_{j=1}^8 w_{ij} = 1$.
We regularize cage coefficients of the target mesh to follow those of the auxiliary mesh by minimizing the following loss function:
\begin{equation}
    \mathcal{L}_{\mathrm{cage}} = \frac{1}{L} \sum_{l=1}^L \frac{1}{|\mathcal{C}_l|} \sum_{\mathbf{v}_i \in \mathcal{C}_l} \| \mathbf{W}_i^{\mathrm{aux}} - \mathbf{W}_i^{\mathrm{tgt}} \|^2_2,
    \label{eq:cage_loss}
\end{equation}
where $|\mathcal{C}_l|$ is the number of vertices in part $\mathcal{C}_l$, and $\mathbf{W}_i^{\mathrm{aux}}$, $\mathbf{W}_i^{\mathrm{tgt}}$ are cage coefficients calculated using the updated OBBs of the auxiliary mesh and target mesh, respectively.
The auxiliary mesh and cage coefficient regularization enable stable and semantically coherent deformations under SDS optimization.

\subsubsection{Symmetry Regularization}
\label{subsec:sym_loss}

We allow users to optionally enforce symmetry during optimization if the source mesh exhibits internal symmetry.
We detect reflectional symmetry based on the vertices of the source mesh. 
We apply PCA to the vertices $\{\mathbf{v}_i \}_{i=1}^V$ and obtain the principal axes $\{\mathbf{a}_k \}_{k=1}^3$. 
For each axis $\mathbf{a}_k$, we define a reflection plane $\mathbf{\Pi}_k$ with normal $\mathbf{a}_k$ passing through the centroid $\bar{\mathbf{v}} = \frac{1}{V} \sum_i \mathbf{v}_i$. 
Then each vertex $\mathbf{v}_i$ is mirrored across $\mathbf{\Pi}_k$ onto $\mathbf{v}_{\mathrm{mir}(i, k)}$, and its nearest vertex $\mathbf{v}_{j(i, k)}$ is identified.
We consider $\mathbf{\Pi}_k$ to be a valid symmetry plane when the following two conditions hold:
\begin{align}
    \| \mathbf{v}_{\mathrm{mir}(i, k)} - \mathbf{v}_{j(i, k)} \|_2 < \tau_1, \forall i \quad \text{and}  \nonumber \\  \sum_{i} \| \mathbf{v}_{\mathrm{mir}(i, k)} - \mathbf{v}_{j(i, k)} \|_2  < \tau_2,
\end{align}
where $\tau_1$ and $\tau_2$ are threshold values. 
We denote the set containing the symmetric pairs $\{ (i, j(i, k)) \}$ as $\mathcal{P}_k$.

Once the symmetry is detected, we introduce a symmetry loss consisting of two regularization terms. 
For each symmetric pair $(i, j(i, k)) \in \mathcal{P}_k$, we force the midpoint of the symmetric pair to lie on a common plane $\tilde{\mathbf{\Pi}}_k$ (which can be changed from the initial symmetric plane).  
The first loss term penalizes the deviation of midpoints from this plane:
\begin{equation}
    \mathcal{L}_{\mathrm{mid}} = \sum_{k} \frac{1}{|\mathcal{P}_k|} \sum_{(i, j(i)) \in \mathcal{P}_k} | \tilde{\mathbf{n}}_k^{\top} (\mathbf{m}_{i, k}- \bar{\mathbf{v}}) |^2.
    \label{eq:sym_loss_1}
\end{equation}
Using the calculated midpoints $\mathbf{m}_{i, k} = \frac{1}{2} ( \mathbf{v}_i + \mathbf{v}_{j(i, k)})$, we calculate a normal vector $\tilde{\mathbf{n}}_k$ of common plane $\tilde{\mathbf{\Pi}}_k$ to all midpoints by performing SVD on their covariance matrix.
The second term encourages the direction vector between the symmetric pair, $\mathbf{d}_{i, k} = \mathbf{v}_i - \mathbf{v}_{j(i, k)}$, to be orthogonal to $\tilde{\mathbf{n}}_k$ by minimizing
\begin{equation}
    \mathcal{L}_{\mathrm{dir}} = \sum_{k} \frac{1}{|\mathcal{P}_k|} \sum_{(i, j(i)) \in \mathcal{P}_k} (1 - |\tilde{\mathbf{n}}_k^{\top} \hat{\mathbf{d}}_{i, k} |),
    \label{eq;sym_loss_2}
\end{equation}
where $ \hat{\mathbf{d}}_{i, k} =  {\mathbf{d}}_{i, k} / \| {\mathbf{d}}_{i, k} \|$.
The complete symmetry loss is given by: $\mathcal{L}_{\mathrm{sym}} = \mathcal{L}_{\mathrm{mid}} + \mathcal{L}_{\mathrm{dir}}$.
Note that symmetry loss is also defined for the auxiliary mesh by treating the centers of spheres as $\{ \mathbf{v}_i \}$, and is denoted as $\mathcal{L}_{\mathrm{sym}}^{\mathrm{aux}}$.

\subsection{Coarse-to-Fine Deformation Pipeline}
\label{subsec:deformation}

In the coarse stage, we optimize the scale, rotation, and translation parameters of the cages of the auxiliary mesh using the following loss function:
\begin{equation}
    \mathcal{L}_{\mathrm{aux}} = \lambda_1 \mathcal{L}_{\mathrm{SDS}}^{\mathrm{aux}} +   \lambda_2 \mathcal{L}_{\mathrm{sym}}^{\mathrm{aux}},
    \label{eq:coarse_loss_aux}
\end{equation}
where $\lambda_1$ and $\lambda_2$ are hyperparameters. 
With the optimized cages, we calculate $\mathcal{L}_{\mathrm{cage}}$ by following Eq.~\eqref{eq:cage_loss}.
Then, the target mesh is optimized with Eq.~\eqref{eq:coarse_loss_tgt}:
\begin{equation}
    \mathcal{L}_{\mathrm{tgt}} (t) = \lambda_3 \mathcal{L}_{\mathrm{SDS}} + \lambda_4 \mathcal{L}_{\mathrm{reg}} +  \lambda_5 \mathcal{L}_{\mathrm{sym}} + \lambda_6(t) \mathcal{L}_{\mathrm{cage}},
    \label{eq:coarse_loss_tgt}
\end{equation}
where $t \in (0, N_1]$ denotes the optimization iteration, and $\lambda_3, \lambda_4, \lambda_5$ are constant hyperparameters.
We linearly decay $\lambda_6(t)$ from its initial value $\lambda_6$ as follows:
\begin{equation}
    \lambda_6(t) = \lambda_6 \left(1 - 0.99 {t}\big/{N_1} \right).
    \label{eq:cage_loss_weight_decay}
\end{equation}
Here, $ \mathcal{L}_{\mathrm{reg}}$ is a Jacobian regularization term introduced in TextDeformer~\citep{gao2023textdeformer} that prevents the deformed mesh from deviating excessively from the source mesh by encouraging  $\{\mathbf{J}_i\}$ to follow the identity matrix.
In the fine stage, we do not regularize cage coefficients and minimize:
\begin{equation}
    \mathcal{L}_{\mathrm{tgt}} = \lambda_3 \mathcal{L}_{\mathrm{SDS}} + \lambda_4 \mathcal{L}_{\mathrm{reg}} +  \lambda_5 \mathcal{L}_{\mathrm{sym}},
    \label{eq:fine_loss_tgt}
\end{equation}
where $t \in (N_1, N_2]$.
This stage applies fine-grained adjustments to capture the geometric style of the reference image while preserving the large-scale translations established in the coarse stage.

\section{Experiments}
\label{sec:experiments}

\begin{figure*}
	\centering
    \includegraphics[width=\linewidth]{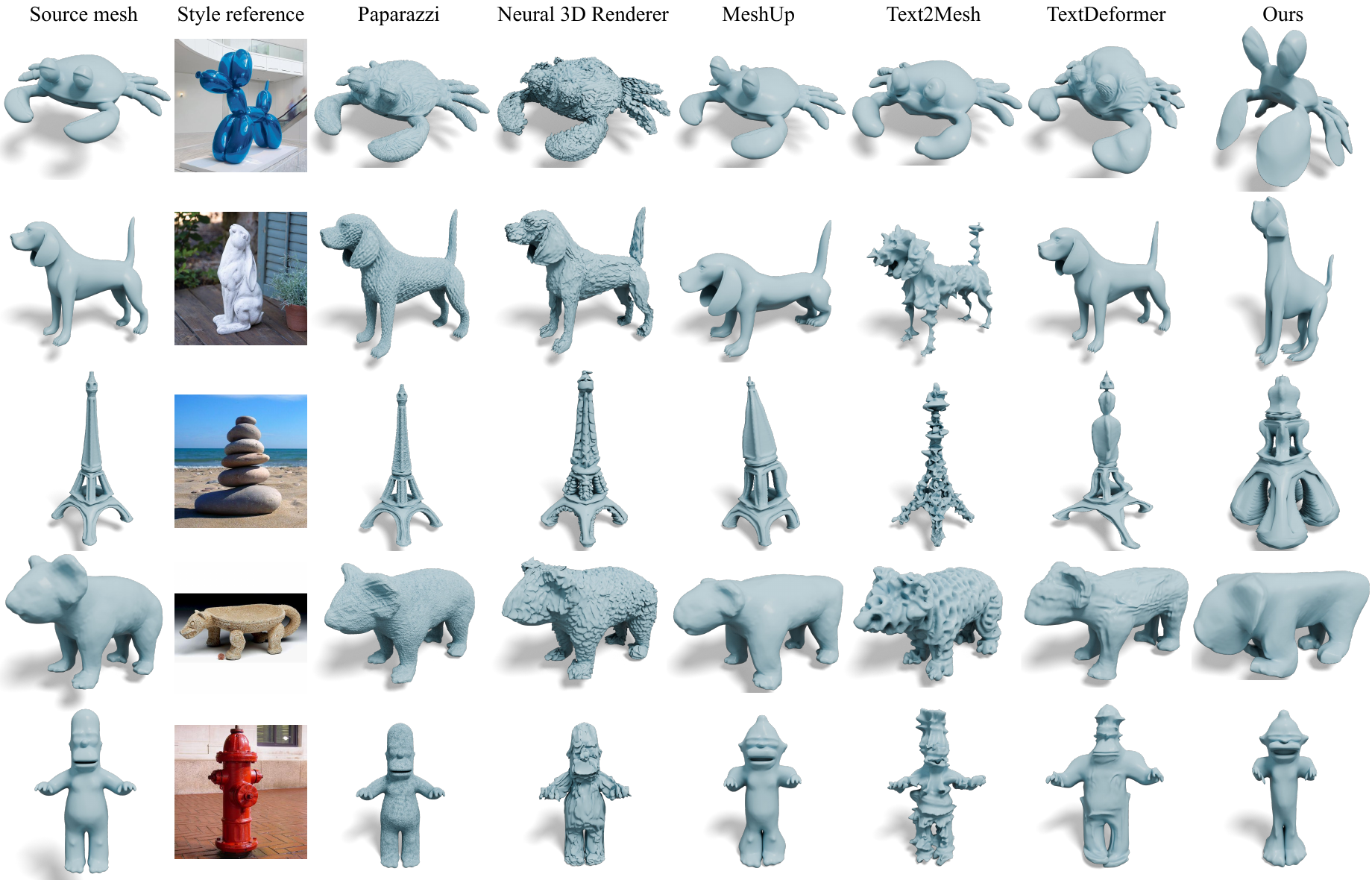}
	\caption{Qualitative comparison of the proposed method against baselines~\cite{liu2018paparazzi, kato2018neural, kim2025meshup, michel2022text2mesh, gao2023textdeformer}.
    Our method achieves expressive geometric deformation, accurately reflecting both coarse structure and fine detail from the style reference while preserving the identity of source mesh, whereas baselines struggle to capture the intended geometry.
    } 
\label{fig:qual_cmp}
\end{figure*}

In this section, we first demonstrate the superiority of our method over baselines through a user study and qualitative comparisons. 
We further highlight the importance of the proposed components via ablation studies.
Finally, we show that our approach flexibly incorporates additional conditioning signals such as texts and user-selected parts.

\subsection{Implementation Details}
\label{subsec:imple}
We train LoRA~\cite{hu2022lora} module of rank 16 with DreamBooth~\cite{ruiz2023dreambooth} using 4-12 reference images.
The source meshes used in our experiments typically contain 2k-20k vertices.
During optimization, we render meshes with differentiable rasterizer~\cite{Laine2020diffrast}. 
Users can adaptively select the number of semantic parts segmented by PartField~\cite{liu2025partfield} for cage coefficient regularization.
Details of the experimental setup are provided in the Appendix A.

\begin{figure}[t!]
	\centering
	\includegraphics[width=1.0\linewidth]{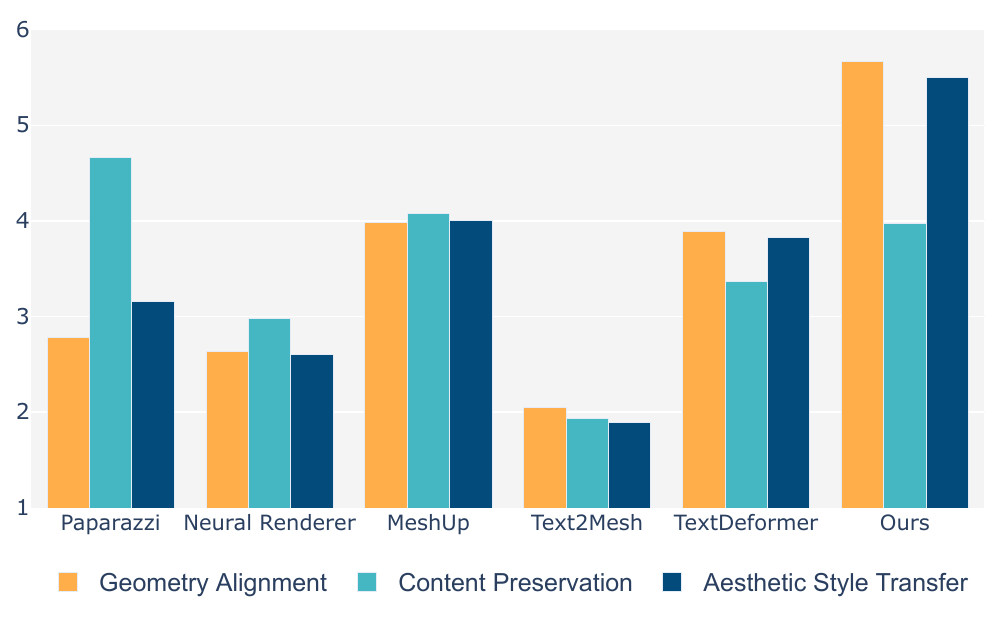}
    \vspace{-4mm}
	\caption{User study results.
    We examine each method using three criteria, including the measurement of geometry alignment, content preservation, and aesthetic style transfer. 
    } 
    \vspace{-4mm}
\label{fig:quant_user_study}
\end{figure}

\begin{figure*}[t!]
	\centering
	\includegraphics[width=\linewidth]{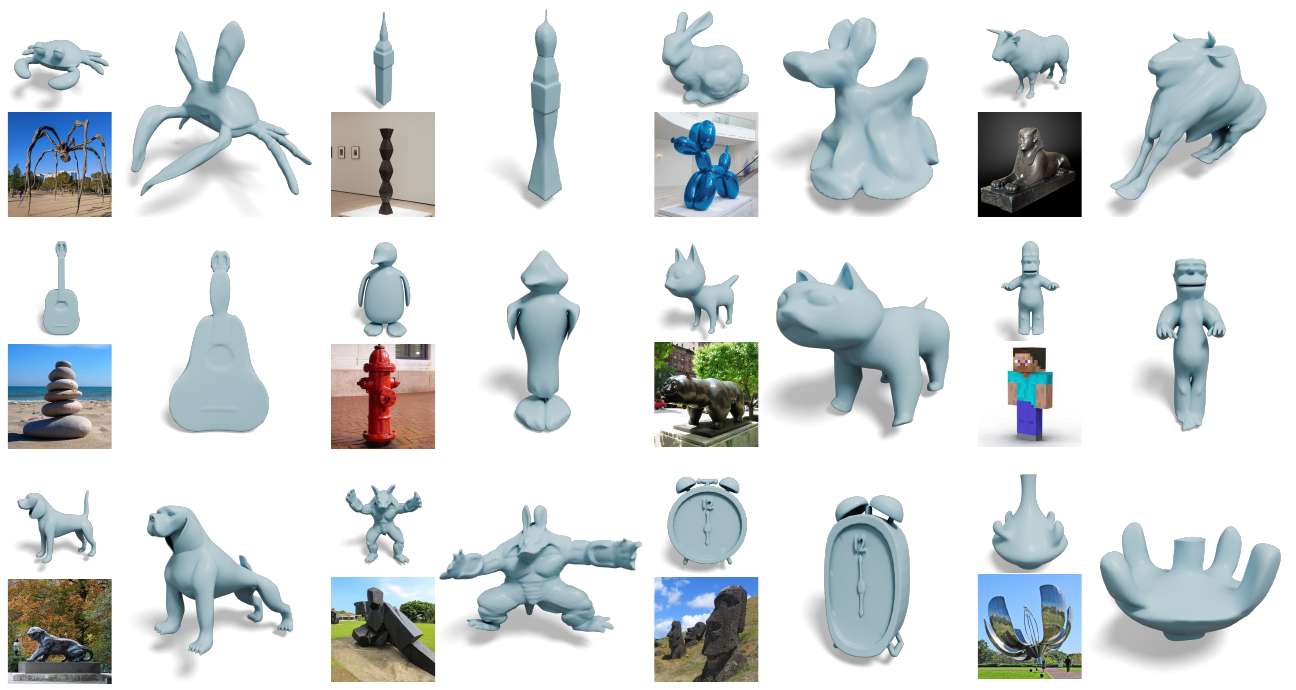}
	\caption{Qualitative geometric stylization results using diverse source mesh and style references.
    Our method effectively transfers the geometric style from the reference image while preserving the overall semantics of the source mesh.
    } 
\label{fig:qual_ours}
\end{figure*}

\subsection{Comparative Evaluation}
\label{subsec:exp_eval}

\paragraph{Quantitative evaluation.}
We compare our method against four baselines: Paparazzi~\cite{liu2018paparazzi}, Neural 3D Mesh Renderer~\cite{kato2018neural}, MeshUp~\cite{kim2025meshup}, Text2Mesh~\cite{michel2022text2mesh}, and TextDeformer~\cite{gao2023textdeformer}. 
Since our task focuses on geometric stylization from image references rather than text prompts, we adapt Text2Mesh and TextDeformer by replacing their CLIP~\cite{radford2021learning} text embeddings with CLIP image embeddings of the reference style references.
We compute the loss with the same set of reference images as those used during LoRA~\cite{hu2022lora} training, and we use the averaged loss value for optimization. 
For MeshUp, we leverage the Textual Inversion~\cite{gal2022image} technique to obtain an optimized token, then use it with the pretrained DeepFloyd-IF~\cite{DeepFloydIF} for deformation.
Then we quantitatively evaluate the proposed method with baselines through a perceptual user study.
A total of 32 participants were gathered, and each participant was requested to rank the outputs of each method for 8 samples based on three criteria:
(1) how well the geometry aligns with the style reference, 
(2) how faithfully the content of the source mesh is preserved,
and (3) how effectively the aesthetic style is transferred.
For every sample, we converted ranks into scores in descending order.
We present the details of user study in Appendix B.
As shown in~\cref{fig:quant_user_study}, our method achieves the best perceptual ranking in terms of geometric alignment and aesthetic style transfer.
We note that baselines may score higher in content preservation since some of them struggle to produce geometric structural changes, resulting in meshes that remain close to the source mesh without reflecting the desired deformation.

\paragraph{Qualitative evaluation.}
We visualize the qualitative comparisons in \cref{fig:qual_cmp}. 
Paparazzi and Neural 3D Mesh Renderer primarily perform texture-oriented style transfer, and therefore struggle to induce a desired geometric deformation, resulting in only local shape variations.
Text2Mesh tends to produce noisy artifacts due to its direct optimization over vertex coordinates and color rather than structured per-face Jacobians.
TextDeformer, which uses Jacobian-based deformation, still fails to capture complex geometric styles, highlighting the limited capability of CLIP-based guidance.
MeshUp, while using the SDS~\cite{poole2022dreamfusion} loss, relies on a pixel-level diffusion model rather than SDXL~\cite{podell2023sdxl} and does not incorporate the  sophisticated techniques used in our method, thus still yielding suboptimal results.
In contrast, our method effectively transforms the global structure, pose, and geometry of the style reference while maintaining the semantic content of the source mesh.
We visualize additional results in \cref{fig:teaser,fig:qual_ours} and Appendix C.

\begin{figure*}
	\centering
	\includegraphics[width=1.0\linewidth]{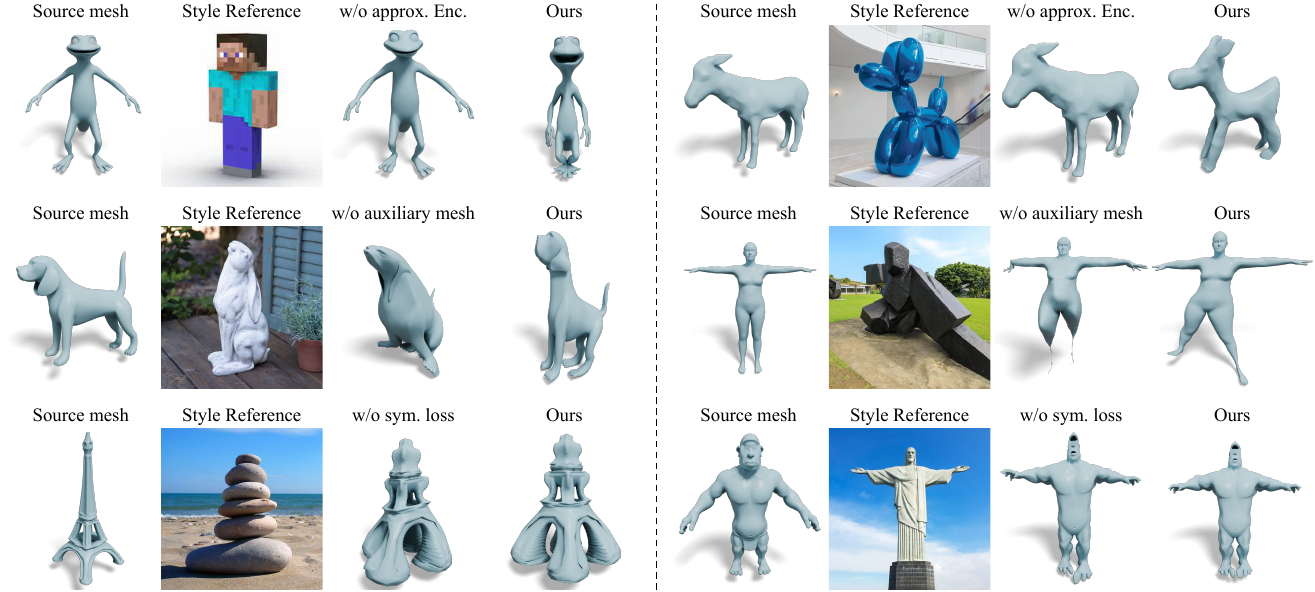}
	\caption{Ablation study results.
    We analyze the effects of the approximated VAE encoder (1st row), cage-coefficient regularization (2nd) and symmetry loss (3rd).
    The qualitative results show that each component is essential for producing high-fidelity stylization result.}
\label{fig:qual_ablation}
\end{figure*}

\subsection{Ablation Study}
\label{subsec:exp_ablation}
We validate the effectiveness of the proposed components through ablation studies, with qualitative results in \cref{fig:qual_ablation}.
Firstly, we examine the role of the approximated VAE~\cite{kingma2013auto} encoder of our image-guided geometric deformation. 
As shown in the first row, using the original SDXL VAE~\cite{podell2023sdxl} fails to produce plausible transforms and tends to remain close to the source mesh, indicating that the approximated VAE encoder is crucial for inducing meaningful shape deformation.
Second, we evaluate the impact of cage coefficient regularization.
As visualized in the second row, without this regularization, the deformed meshes often exhibit distorted geometry and fail to accurately reflect the geometric style encoded from the reference image.  
These results show that the auxiliary mesh-based regularization not only facilitates large geometry transforms, but also stabilizes the overall deformation process.
Lastly, we analyze the symmetry loss.
As shown in the final row, enforcing symmetry is beneficial when the source mesh and the intended translation inherently possess symmetric structures.
This optional constraint allows users to achieve more visually consistent deformations in such scenarios.

\subsection{Additional Results}
\label{subsec:exp_additional_result}

\begin{figure}
	\centering
	\includegraphics[width=1.0\linewidth]{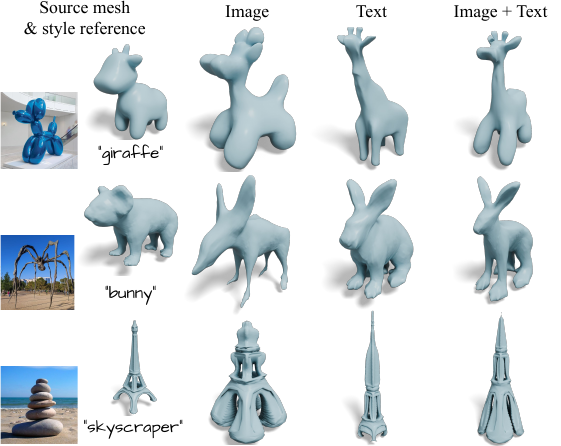}
	\caption{Geometric stylization results with text conditioning. 
    Our method effectively incorporates textual prompts alongside style reference images.
    The resulting meshes align with the text-described contents while consistently maintaining the geometric style encoded in the style reference.
    }  
\label{fig:qual_text_img}
\end{figure}

\paragraph{Geometric stylization with text conditions.}
We further demonstrate that our method can be flexibly combined with additional conditions.
First, we use text prompts as control signals, showing that our method enables simultaneous content manipulation and style transfer.
In this setting, the source meshes are deformed based on both text instructions and style references.
For instance, as illustrated in the first row of \cref{fig:qual_text_img}, the proposed method transforms the source mesh into a giraffe while transferring the overall style of the given sculpture image.

\begin{figure}
	\centering
	\includegraphics[width=\linewidth]{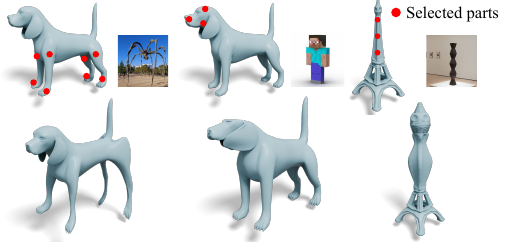}
	\caption{Our approach can transfer geometric style only to user-selected regions while preserving the original shape elsewhere.
    }  
    \vspace{-1em}
\label{fig:qual_local_condition}
\end{figure}

\paragraph{Localized deformations.}
We further demonstrate that the proposed method can perform localized geometric stylization.
In this setting, users specify one or more regions of the source mesh to be deformed.
In practice, we adopt the part segmentation defined by PartField~\cite{liu2025partfield}, and visualize the selected regions as point sets in \cref{fig:qual_local_condition}.
During optimization, we compute the target loss $\mathcal{L}_{\mathrm{tgt}}$ and backpropagate gradients only through the Jacobians corresponding to the selected parts.
As shown, the proposed method enables flexible and targeted style transfer, allowing stylization of either the entire mesh or localized areas while preserving the geometry elsewhere.

\section{Conclusion}
\label{sec:conclusion}

In this work, we propose a novel framework for geometric stylization of 3D meshes. 
Unlike prior approaches that primarily focus on texture-oriented stylization or text-based mesh deformation, our method emphasizes geometric style and derives it directly from reference images.
We extract style information by training LoRA on a set of style images via DreamBooth, and apply it to the source mesh through a Jacobian-based deformation guided by SDS loss.
To further improve the expressiveness of the deformation and computational efficiency, we adopt Stable Diffusion XL with an approximated VAE encoder. 
We also introduce both cage coefficient regularization and a symmetry loss to enable large-scale geometry manipulations while preserving the inherent structural symmetries.
Experimental results demonstrate that our method produces semantically consistent geometric stylizations, outperforming existing baselines.

\section*{Acknowledgements}

This work was supported by the National Research Foundation of Korea(NRF) grant (No. RS-2026-25485899) and the Institute of Information \& Communications Technology Planning \& Evaluation(IITP) grant (RS-2025-25442338, AI star Fellowship Support Program(Seoul National Univ.)) funded by the Korea government(MSIT).

{
    \small
    \bibliographystyle{ieeenat_fullname}
    \bibliography{references}

@String(CVPR= {IEEE Conf. Comput. Vis. Pattern Recog.})

@String(ICCV= {Int. Conf. Comput. Vis.})

@String(ECCV= {Eur. Conf. Comput. Vis.})

@String(NIPS= {Adv. Neural Inform. Process. Syst.})

@String(TOG= {ACM Trans. Graph.})

@String(ICLR = {Int. Conf. Learn. Represent.})

@String(CVPR  = {CVPR})

@String(ICCV  = {ICCV})

@String(ECCV  = {ECCV})

@String(NIPS  = {NeurIPS})

@String(TOG   = {ACM TOG})

@String(ICLR  = {ICLR})

@article{aigerman2022neural,
  title={Neural jacobian fields: Learning intrinsic mappings of arbitrary meshes},
  author={Aigerman, Noam and Gupta, Kunal and Kim, Vladimir G and Chaudhuri, Siddhartha and Saito, Jun and Groueix, Thibault},
  journal={SIGGRAPH},
  year={2022}
}

@inproceedings{gao2023textdeformer,
  title={Textdeformer: Geometry manipulation using text guidance},
  author={Gao, William and Aigerman, Noam and Groueix, Thibault and Kim, Vova and Hanocka, Rana},
  booktitle={SIGGRAPH},
  year={2023}
}

@inproceedings{ruiz2023dreambooth,
  title={Dreambooth: Fine tuning text-to-image diffusion models for subject-driven generation},
  author={Ruiz, Nataniel and Li, Yuanzhen and Jampani, Varun and Pritch, Yael and Rubinstein, Michael and Aberman, Kfir},
  booktitle={CVPR},
  year={2023}
}

@article{ho2020denoising,
  title={Denoising diffusion probabilistic models},
  author={Ho, Jonathan and Jain, Ajay and Abbeel, Pieter},
  journal={NeurIPS},
  year={2020}
}

@article{song2020denoising,
  title={Denoising diffusion implicit models},
  author={Song, Jiaming and Meng, Chenlin and Ermon, Stefano},
  journal={ICLR},
  year={2021}
}

@article{hu2022lora,
  title={Lora: Low-rank adaptation of large language models.},
  author={Hu, Edward J and Shen, Yelong and Wallis, Phillip and Allen-Zhu, Zeyuan and Li, Yuanzhi and Wang, Shean and Wang, Lu and Chen, Weizhu and others},
  journal={ICLR},
  year={2022}
}

@article{podell2023sdxl,
  title={Sdxl: Improving latent diffusion models for high-resolution image synthesis},
  author={Podell, Dustin and English, Zion and Lacey, Kyle and Blattmann, Andreas and Dockhorn, Tim and M{\"u}ller, Jonas and Penna, Joe and Rombach, Robin},
  journal={arXiv:2307.01952},
  year={2023}
}

@inproceedings{ronneberger2015u,
  title={U-net: Convolutional networks for biomedical image segmentation},
  author={Ronneberger, Olaf and Fischer, Philipp and Brox, Thomas},
  booktitle={International Conference on Medical image computing and computer-assisted intervention},
  year={2015},
}

@inproceedings{rombach2022high,
  title={High-resolution image synthesis with latent diffusion models},
  author={Rombach, Robin and Blattmann, Andreas and Lorenz, Dominik and Esser, Patrick and Ommer, Bj{\"o}rn},
  booktitle={CVPR},
    year={2022}
}

@article{poole2022dreamfusion,
  title={Dreamfusion: Text-to-3d using 2d diffusion},
  author={Poole, Ben and Jain, Ajay and Barron, Jonathan T and Mildenhall, Ben},
  journal={ICLR},
  year={2023}
}

@article{kingma2013auto,
  title={Auto-encoding variational bayes},
  author={Kingma, Diederik P and Welling, Max},
  journal={arXiv:1312.6114},
  year={2013}
}

@misc{DeepFloydIF,
author={DeepFloyd Lab at StabilityAI},
title={{DeepFloyd IF}: a novel state-of-the-art open-source text-to-image model with a high degree of photorealism and language understanding},
howpublished="\url{https://www.deepfloyd.ai/deepfloyd-if}",
year={2023},
}

@inproceedings{kim2025meshup,
  title={Meshup: Multi-target mesh deformation via blended score distillation},
  author={Kim, Hyunwoo and Lang, Itai and Aigerman, Noam and Groueix, Thibault and Kim, Vladimir G and Hanocka, Rana},
  booktitle={3DV},
  year={2025},
}

@article{liu2025partfield,
  title={Partfield: Learning 3d feature fields for part segmentation and beyond},
  author={Liu, Minghua and Uy, Mikaela Angelina and Xiang, Donglai and Su, Hao and Fidler, Sanja and Sharp, Nicholas and Gao, Jun},
  journal={ICCV},
  year={2025}
}

@article{Laine2020diffrast,
  title   = {Modular Primitives for High-Performance Differentiable Rendering},
  author  = {Samuli Laine and Janne Hellsten and Tero Karras and Yeongho Seol and Jaakko Lehtinen and Timo Aila},
  journal = {ACM TOG},
  year    = {2020},
}

@article{liu2018paparazzi,
  title={Paparazzi: surface editing by way of multi-view image processing.},
  author={Liu, Hsueh-Ti Derek and Tao, Michael and Jacobson, Alec},
  journal={ACM TOG},
  year={2018}
}

@inproceedings{kato2018neural,
  title={Neural 3d mesh renderer},
  author={Kato, Hiroharu and Ushiku, Yoshitaka and Harada, Tatsuya},
  booktitle={CVPR},
  year={2018}
}

@inproceedings{michel2022text2mesh,
  title={Text2mesh: Text-driven neural stylization for meshes},
  author={Michel, Oscar and Bar-On, Roi and Liu, Richard and Benaim, Sagie and Hanocka, Rana},
  booktitle={CVPR},
  year={2022}
}

@inproceedings{radford2021learning,
  title={Learning transferable visual models from natural language supervision},
  author={Radford, Alec and Kim, Jong Wook and Hallacy, Chris and Ramesh, Aditya and Goh, Gabriel and Agarwal, Sandhini and Sastry, Girish and Askell, Amanda and Mishkin, Pamela and Clark, Jack and others},
  booktitle={ICML},
  year={2021}
}

@inproceedings{gatys2016image,
  title={Image style transfer using convolutional neural networks},
  author={Gatys, Leon A and Ecker, Alexander S and Bethge, Matthias},
  booktitle={CVPR},
  year={2016}
}

@article{chen2016fast,
  title={Fast patch-based style transfer of arbitrary style},
  author={Chen, Tian Qi and Schmidt, Mark},
  journal={arXiv:1612.04337},
  year={2016}
}

@inproceedings{kolkin2019style,
  title={Style transfer by relaxed optimal transport and self-similarity},
  author={Kolkin, Nicholas and Salavon, Jason and Shakhnarovich, Gregory},
  booktitle={CVPR},
  year={2019}
}

@inproceedings{li2016combining,
  title={Combining markov random fields and convolutional neural networks for image synthesis},
  author={Li, Chuan and Wand, Michael},
  booktitle={CVPR},
  year={2016}
}

@inproceedings{gu2018arbitrary,
  title={Arbitrary style transfer with deep feature reshuffle},
  author={Gu, Shuyang and Chen, Congliang and Liao, Jing and Yuan, Lu},
  booktitle={CVPR},
  year={2018}
}

@article{liao2017visual,
  title={Visual attribute transfer through deep image analogy},
  author={Liao, Jing and Yao, Yuan and Yuan, Lu and Hua, Gang and Kang, Sing Bing},
  journal={arXiv:1705.01088},
  year={2017}
}

@inproceedings{mechrez2018contextual,
  title={The contextual loss for image transformation with non-aligned data},
  author={Mechrez, Roey and Talmi, Itamar and Zelnik-Manor, Lihi},
  booktitle={ECCV},
  year={2018}
}

@article{risser2017stable,
  title={Stable and controllable neural texture synthesis and style transfer using histogram losses},
  author={Risser, Eric and Wilmot, Pierre and Barnes, Connelly},
  journal={arXiv:1701.08893},
  year={2017}
}

@inproceedings{kotovenko2021rethinking,
  title={Rethinking style transfer: From pixels to parameterized brushstrokes},
  author={Kotovenko, Dmytro and Wright, Matthias and Heimbrecht, Arthur and Ommer, Bjorn},
  booktitle={CVPR},
  year={2021}
}

@inproceedings{kotovenko2019content_cvpr,
  title={A content transformation block for image style transfer},
  author={Kotovenko, Dmytro and Sanakoyeu, Artsiom and Ma, Pingchuan and Lang, Sabine and Ommer, Bjorn},
  booktitle={CVPR},
  year={2019}
}

@inproceedings{kotovenko2019content,
  title={Content and style disentanglement for artistic style transfer},
  author={Kotovenko, Dmytro and Sanakoyeu, Artsiom and Lang, Sabine and Ommer, Bjorn},
  booktitle={ICCV},
  year={2019}
}

@inproceedings{huang2017arbitrary,
  title={Arbitrary style transfer in real-time with adaptive instance normalization},
  author={Huang, Xun and Belongie, Serge},
  booktitle={ICCV},
  year={2017}
}

@inproceedings{an2021artflow,
  title={Artflow: Unbiased image style transfer via reversible neural flows},
  author={An, Jie and Huang, Siyu and Song, Yibing and Dou, Dejing and Liu, Wei and Luo, Jiebo},
  booktitle={CVPR},
  year={2021}
}

@inproceedings{park2019arbitrary,
  title={Arbitrary style transfer with style-attentional networks},
  author={Park, Dae Young and Lee, Kwang Hee},
  booktitle={CVPR},
  year={2019}
}

@inproceedings{sheng2018avatar,
  title={Avatar-net: Multi-scale zero-shot style transfer by feature decoration},
  author={Sheng, Lu and Lin, Ziyi and Shao, Jing and Wang, Xiaogang},
  booktitle={CVPR},
  year={2018}
}

@article{li2017universal,
  title={Universal style transfer via feature transforms},
  author={Li, Yijun and Fang, Chen and Yang, Jimei and Wang, Zhaowen and Lu, Xin and Yang, Ming-Hsuan},
  journal={NIPS},
  year={2017}
}

@inproceedings{huang2021learning,
  title={Learning to stylize novel views},
  author={Huang, Hsin-Ping and Tseng, Hung-Yu and Saini, Saurabh and Singh, Maneesh and Yang, Ming-Hsuan},
  booktitle={ICCV},
  year={2021}
}

@inproceedings{mu20223d,
  title={3d photo stylization: Learning to generate stylized novel views from a single image},
  author={Mu, Fangzhou and Wang, Jian and Wu, Yicheng and Li, Yin},
  booktitle={CVPR},
  year={2022}
}

@inproceedings{hollein2022stylemesh,
  title={Stylemesh: Style transfer for indoor 3d scene reconstructions},
  author={H{\"o}llein, Lukas and Johnson, Justin and Nie{\ss}ner, Matthias},
  booktitle={CVPR},
  year={2022}
}

@inproceedings{zhang2022arf,
  title={Arf: Artistic radiance fields},
  author={Zhang, Kai and Kolkin, Nick and Bi, Sai and Luan, Fujun and Xu, Zexiang and Shechtman, Eli and Snavely, Noah},
  booktitle={ECCV},
  year={2022},
}

@inproceedings{zhang2023ref,
  title={Ref-npr: Reference-based non-photorealistic radiance fields for controllable scene stylization},
  author={Zhang, Yuechen and He, Zexin and Xing, Jinbo and Yao, Xufeng and Jia, Jiaya},
  booktitle={CVPR},
  year={2023}
}

@inproceedings{pang2023locally,
  title={Locally stylized neural radiance fields},
  author={Pang, Hong-Wing and Hua, Binh-Son and Yeung, Sai-Kit},
  booktitle={ICCV},
  year={2023},
}

@inproceedings{chiang2022stylizing,
  title={Stylizing 3d scene via implicit representation and hypernetwork},
  author={Chiang, Pei-Ze and Tsai, Meng-Shiun and Tseng, Hung-Yu and Lai, Wei-Sheng and Chiu, Wei-Chen},
  booktitle={WACV},
  year={2022}
}

@inproceedings{jung2024geometry,
  title={Geometry transfer for stylizing radiance fields},
  author={Jung, Hyunyoung and Nam, Seonghyeon and Sarafianos, Nikolaos and Yoo, Sungjoo and Sorkine-Hornung, Alexander and Ranjan, Rakesh},
  booktitle={CVPR},
  year={2024}
}

@incollection{liu2024stylegaussian,
  title={Stylegaussian: Instant 3d style transfer with gaussian splatting},
  author={Liu, Kunhao and Zhan, Fangneng and Xu, Muyu and Theobalt, Christian and Shao, Ling and Lu, Shijian},
  booktitle={SIGGRAPH Asia Technical Communications},
  year={2024}
}

@article{zhang2024stylizedgs,
  title={Stylizedgs: Controllable stylization for 3d gaussian splatting},
  author={Zhang, Dingxi and Yuan, Yu-Jie and Chen, Zhuoxun and Zhang, Fang-Lue and He, Zhenliang and Shan, Shiguang and Gao, Lin},
  journal={TPAMI},
  year={2025}
}

@article{liu2020geometric,
  title={Geometric style transfer},
  author={Liu, Xiao-Chang and Li, Xuan-Yi and Cheng, Ming-Ming and Hall, Peter},
  journal={arXiv:2007.05471},
  year={2020}
}

@inproceedings{yin20213dstylenet,
  title={3dstylenet: Creating 3d shapes with geometric and texture style variations},
  author={Yin, Kangxue and Gao, Jun and Shugrina, Maria and Khamis, Sameh and Fidler, Sanja},
  booktitle={ICCV},
  year={2021}
}

@inproceedings{vggnet,
  title = {Very deep convolutional networks for large-scale image recognition},
  author = {Simonyan, K and Zisserman, A},
  booktitle = {ICLR},
  year = {2015},
}

@article{cubic_stylization,
  title = {Cubic Stylization},
  author = {Hsueh-Ti Derek Liu and Alec Jacobson},
  year = {2019},
  journal = {ACM TOG}, 
}

@inproceedings{liu2021normal,
  title={Normal-driven spherical shape analogies},
  author={Liu, Hsueh-Ti Derek and Jacobson, Alec},
  booktitle={Computer Graphics Forum},
  year={2021},
}

@article{gauss_stylization,
title = {{Gauss Stylization: Interactive Artistic Mesh Modeling based on Preferred Surface Normals}},
author = {Kohlbrenner, Maximilian and Finnendahl, Ugo and Djuren, Tobias and Alexa, Marc},
journal = {Computer Graphics Forum},
year = {2021}
}

@inproceedings{dinh2025geometry,
  title={Geometry in Style: 3D Stylization via Surface Normal Deformation},
  author={Dinh, Nam Anh and Lang, Itai and Kim, Hyunwoo and Stein, Oded and Hanocka, Rana},
  booktitle={CVPR},
  year={2025}
}

@article{yaniv2019face,
  title={The face of art: landmark detection and geometric style in portraits},
  author={Yaniv, Jordan and Newman, Yael and Shamir, Ariel},
  journal={ACM TOG},
  year={2019},
}

@inproceedings{dynamic_mesh_stylization,
title = {Controllable Neural Style Transfer for Dynamic Meshes},
author = {Gomes Haetinger, Guilherme and Tang, Jingwei and Ortiz, Raphael and Kanyuk, Paul and Azevedo, Vinicius},
booktitle = {SIGGRAPH},
year = {2024},
}

@article{kolkin2022neural,
  title={Neural neighbor style transfer},
  author={Kolkin, Nicholas and Kucera, Michal and Paris, Sylvain and Sykora, Daniel and Shechtman, Eli and Shakhnarovich, Greg},
  journal={arXiv:2203.13215},
  year={2022}
}

@article{geometric_texture,
author = {Hertz, Amir and Hanocka, Rana and Giryes, Raja and Cohen-Or, Daniel},
title = {Deep geometric texture synthesis},
year = {2020},
journal = {ACM TOG},
}

@inproceedings{metzer2023latent,
  title={Latent-nerf for shape-guided generation of 3d shapes and textures},
  author={Metzer, Gal and Richardson, Elad and Patashnik, Or and Giryes, Raja and Cohen-Or, Daniel},
  booktitle={CVPR},
  year={2023}
}

@article{gal2022image,
  title={An image is worth one word: Personalizing text-to-image generation using textual inversion},
  author={Gal, Rinon and Alaluf, Yuval and Atzmon, Yuval and Patashnik, Or and Bermano, Amit H and Chechik, Gal and Cohen-Or, Daniel},
  journal={arXiv:2208.01618},
  year={2022}
}

@article{heusel2017gans,
  title={Gans trained by a two time-scale update rule converge to a local nash equilibrium},
  author={Heusel, Martin and Ramsauer, Hubert and Unterthiner, Thomas and Nessler, Bernhard and Hochreiter, Sepp},
  journal={NIPS},
  year={2017}
}
}

\clearpage
\maketitlesupplementary
\appendix

\section{Implementation Details}
\label{sec:supp_implementation}

To train LoRA weight~\cite{hu2022lora} on Stable Diffusion XL~\cite{podell2023sdxl} using DreamBooth~\cite{ruiz2023dreambooth}, we use a total of 4-12 images for each style reference and set the LoRA rank to 16.
Following common practice, we adopt the prompt ``A photo of TOK sculpture'' and omit the class-specific prior preservation loss proposed in the original DreamBooth formulation.
The learning rate is set to $10^{-4}$, and LoRA weight is trained for a total of 800 iterations.
In the case of the approximated VAE~\cite{kingma2013auto} encoder, we first render $N=500$ images of the source mesh from random viewpoints and obtain corresponding latent via SDXL VAE encoder. 
We then fit the matrix using these 500 image--latent pairs.

For mesh deformation, the SDS loss~\cite{poole2022dreamfusion} is computed using the text prompt ``A TOK style sculpture''.
We use $N_1 = 1800$ and $N_2 = 3600$ iterations for optimization.
Rasterization is performed with the differentiable nvdiffrast rasterizer~\cite{Laine2020diffrast}.
The camera FOV and distance are set to $30\degree$ and 5.0, respectively.
The viewpoint is randomly sampled by choosing the elevation uniformly from [$10\degree$, $30\degree$] and the azimuth from [$0 \degree$, $360 \degree$).
Batch size is set to 4, \textit{i.e} We render mesh from four different viewpoint at each iteration.
We optionally leverage the proposed symmetry loss, when the symmetry is detected via Eq. (8) of the main paper and the source mesh's vertices are arranged accordingly.

\section{Details of User Study}
\label{sec:supp_limitations}

We quantitatively compare the proposed method with baselines through a perceptual user study.
A total of 32 participants were gathered and each participant was requested to rank the output of a total of six different methods~\cite{liu2018paparazzi, kato2018neural, kim2025meshup, michel2022text2mesh, gao2023textdeformer} for 8 samples based on three criteria:
\begin{itemize}
    \item \textbf{1) Geometric Alignment}: Rank the 3D model based on how accurately it aligns with the reference image in terms of geometric properties (pose, silhouette, and structural shapes). (A higher rank indicates closer geometric correspondence to the reference image.)
    \item \textbf{2) Content Preservation}: Rank the 3D model based on how well it maintains characteristics of the original 3D model, even after incorporating the reference image’s style. (A higher rank means the 3D model remains more faithful to the original model while integrating the new style.)
    \item \textbf{3) Aesthetic Style Consistency}: Rank the 3D model based on how well it captures the overall artistic style or visual impression of the reference image. (A higher rank means the model conveys a style that feels more consistent with the reference.)
\end{itemize}
For each question, we randomly shuffle the order of the outputs from six different methods to ensure a fair comparison.

\section{Additional Results}
\label{sec:supp_additional_results}

\begin{figure*}[t!]
	\centering
	\includegraphics[width=1.0\linewidth]{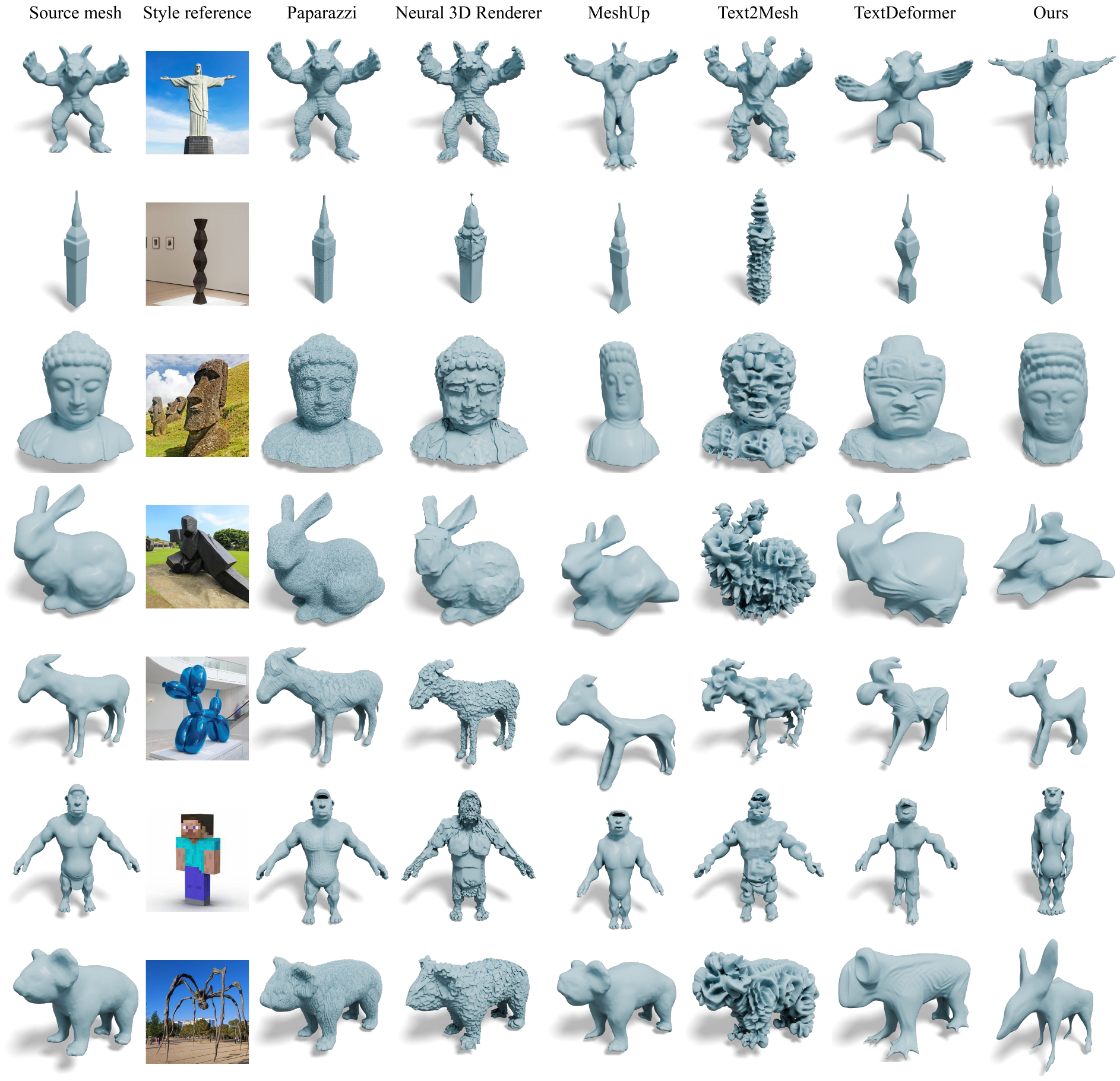}
	\caption{Additional qualitative comparison of our method with the baselines~\cite{liu2018paparazzi, kato2018neural, kim2025meshup, michel2022text2mesh, gao2023textdeformer} on diverse geometric stylization scenarios.
    }
    \label{fig:supp_comp}
\end{figure*}

\begin{figure*}[t!]
	\centering
	\includegraphics[width=1.0\linewidth]{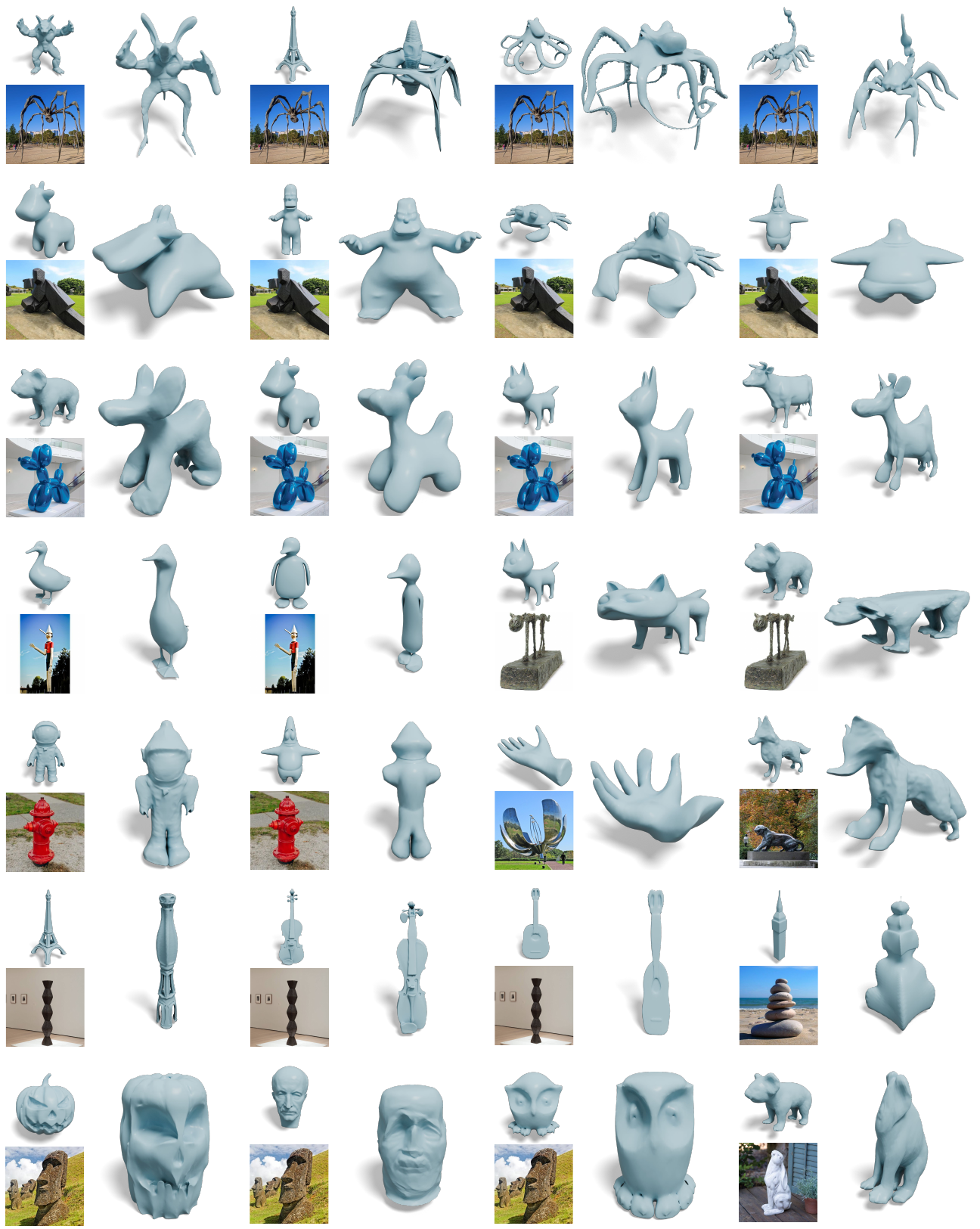}
	\caption{Additional qualitative results of the proposed method.
    }
    \label{fig:supp_qual}
\end{figure*}

We visualize the additional qualitative comparison with the baselines~\cite{liu2018paparazzi, kato2018neural, michel2022text2mesh, gao2023textdeformer} in \cref{fig:supp_comp}.
As shown, our method effectively deforms the source mesh to incorporate the geometric style of the style reference, while baseline methods tend to generate artifacts or only texture-like small modifications.
We show the additional qualitative results in \cref{fig:supp_qual}, highlighting the superior performance of the proposed method.

In addition, we report the FID score~\cite{heusel2017gans} between rendered meshes and style references for additional quantitative evaluation.
Each deformed mesh is rendered from 16 different viewpoints, and the corresponding style images employed during LoRA~\cite{hu2022lora} training are used as the reference set.
A total of 12 meshes are used for evaluation.
As reported in Table~\ref{tab:supp_fid}, our method achieves the best score, demonstrating superior perceptual alignment with the target style compared to baselines.

\vspace{-2mm}
\begin{table}[h!]
\centering
\resizebox{0.65\linewidth}{!}{%
\setlength{\tabcolsep}{7pt}%
\begin{tabular}{@{}cc@{}}
\toprule
Method & FID ($\downarrow$) \\
\midrule
Paparazzi~\cite{liu2018paparazzi} & 419.89 \\
Neural 3D Mesh Renderer~\cite{kato2018neural} & 417.08 \\
MeshUp~\cite{kim2025meshup} & 396.23 \\
Text2Mesh~\cite{michel2022text2mesh} & 409.17 \\
TextDeformer~\cite{gao2023textdeformer} & 400.58 \\
Ours & \textbf{376.57} \\
\bottomrule
\end{tabular}
}
\vspace{-2mm}
\caption{Quantitative comparison measured by FID~\cite{heusel2017gans}.}
\label{tab:supp_fid}
\end{table}

\end{document}